\documentclass{article}

\usepackage[main, final]{neurips_2026}

\usepackage[utf8]{inputenc} 
\usepackage[T1]{fontenc}    
\usepackage{hyperref}       
\usepackage{url}            
\usepackage{booktabs}       
\usepackage{amsfonts}       
\usepackage{nicefrac}       
\usepackage{microtype}      
\usepackage{xcolor}         

\usepackage{algorithm}
\usepackage{enumitem}
\usepackage{graphicx}
\usepackage{amssymb}
\usepackage{amsmath, amsthm}
\usepackage{algpseudocode}  
\usepackage{xcolor}

\definecolor{stageblue}{RGB}{0,102,204}      
\definecolor{commentgreen}{RGB}{0,128,0}     
\definecolor{hintorange}{RGB}{204,102,0}     

\newcommand{\stage}[1]{\textcolor{stageblue}{\textbf{\textit{// #1}}}}

\title{PSD: Pushing the Pareto Frontier of Diffusion LLMs via Parallel Speculative Decoding}

\author{
  Shengyin Sun\thanks{Equal contribution.} \\
  Department of Computer Science\\
  City University of Hong Kong\\
  \texttt{shengysun4-c@my.cityu.edu.hk} \\
  \And
  Yiming Li\footnotemark[1]\,\,\,\thanks{Project lead.} \\
  Huawei Technologies \\
  \texttt{li.yiming3@huawei.com} \\
  \And
  Renxi Liu \\
  Huawei Technologies \\
  \And
  Xinqi Li \\
  Huawei Technologies \\
  \And
  Hui-Ling Zhen \\
  Huawei Technologies \\
  \And
  \textbf{Weizhe Lin} \\
  Huawei Technologies \\
  \And
  \textbf{Chen Chen} \\
  Huawei Technologies \\
  \And
  \textbf{Xianzhi Yu} \\
  Huawei Technologies \\
  \And
  \textbf{Mingxuan Yuan} \\
  Huawei Technologies \\
  \And
  \textbf{Chen Ma}\thanks{Corresponding author.} \\
  Department of Computer Science\\
  City University of Hong Kong\\
  \texttt{chenma@cityu.edu.hk} \\
}

\begin{document}

\maketitle

\begin{abstract}
  Diffusion large language models (dLLMs) generate text by iteratively denoising masked token sequences. Although dLLMs can predict all masked positions in parallel within each step, the large number of denoising iterations still makes inference expensive. This cost can be reduced spatially by unmasking multiple tokens per step, or temporally by collapsing multiple denoising steps into one verification call. We propose Parallel Speculative Decoding (PSD), a training-free framework that jointly improves inference along both axes. Using the confidence scores from a single forward pass, PSD selects positions to unmask via a configurable, adaptive unmasking policy and constructs multi-depth speculative drafts without extra model calls. A final batched verification pass then applies hierarchical acceptance, keeping the deepest draft that remains consistent with the updated predictions. Experiments on three dLLMs across reasoning and code generation tasks show that PSD achieves favorable trade-offs between inference efficiency and generation quality, reaching up to $5.5\times$ tokens per forward pass with accuracy comparable to greedy decoding.

\end{abstract}

\section{Introduction}
\label{sec:intro}

Large language models (LLMs) have demonstrated remarkable capabilities in reasoning and generation~\citep{openai-o1,deepseekr1,DBLP:arxiv/AaronAbhimanyu24}, yet the autoregressive paradigm suffers from sequential token-by-token decoding, leading to high latency and poor hardware utilization~\citep{DBLP:conf/mlsys/PopeDouglas23,DBLP:conf/mlsys/IvanovDryden21,DBLP:arxiv/SunHu25}. Diffusion language models (dLLMs)~\citep{DBLP:arxiv/NieZhu25,DBLP:arxiv/YeXie25,DBLP:authorea/LinJia26} have recently emerged as a promising alternative, formulating text generation as iterative denoising over discrete masked token sequences. Starting from a fully masked sequence, a dLLM progressively reveals tokens over multiple denoising steps, and at each step the model produces predictions for all masked positions simultaneously. This property makes dLLMs naturally amenable to parallel token generation, opening a path toward potentially more efficient inference.

Despite this potential, dLLM inference still requires many denoising iterations, each involving a model forward pass, and deployment efficiency remains a practical bottleneck \citep{DBLP:arxiv/FuWhalen25,DBLP:arxiv/PengLiu25}. Two families of acceleration strategies have been proposed for dLLMs. Parallel decoding methods~\citep{DBLP:arxiv/KongZhang25,DBLP:arxiv/XuJin25,DBLP:conf/iclr/WuZhang26a,DBLP:conf/iclr/WuZhang26b} compress the {spatial} dimension by unmasking multiple tokens per step, but quality degrades as the parallelism grows more aggressive, since low-confidence tokens are forced to commit prematurely and errors propagate through subsequent steps. Speculative decoding methods~\citep{DBLP:arxiv/chenborgeaud23,DBLP:arxiv/AgrawalGarrepalli25} instead compress the {temporal} dimension by anticipating future denoising iterations and verifying them in a single batched forward pass; however, existing approaches still unmask only one token per denoising step, leaving the spatial dimension unexploited and thereby capping the speedup.

Each strategy, however, faces an inherent ceiling when applied in isolation. For parallel decoding, increasing the number of tokens unmasked per step reduces the total number of iterations but inevitably degrades generation quality, yielding a quality--speed Pareto frontier that no transfer policy can escape. For speculative decoding, the speedup is bounded by the acceptance rate of speculated tokens; since each step still reveals only a single token, even perfect speculation yields limited gains. Both families therefore saturate at moderate acceleration levels, leaving considerable room for improvement. We observe, however, that these two axes of acceleration are largely complementary: the spatial axis governs how many tokens are revealed within a single denoising step, while the temporal axis governs how many denoising steps are collapsed into one verification call, and we find empirically that their speedups combine to deliver substantially larger gains than either axis alone. 

In this work, we propose \textit{Parallel Speculative Decoding (PSD)}, a framework that exploits this complementarity by unifying spatial and temporal acceleration for dLLMs (as shown in Figure~\ref{fig:spd_framework}). Rather than prescribing a specific parallel decoding policy, PSD is a general framework that can be built on arbitrary parallel decoding algorithms. PSD builds on a key observation: the set of high-confidence masked positions tends to stabilize as denoising progresses, so the positions currently identified as high-confidence will largely remain high-confidence after a few more tokens are revealed.

More specifically, PSD operates in three stages (Figure~\ref{fig:spd_framework}): (1)~spatial parallel unmasking applies a configurable transfer policy to reveal multiple tokens per step; (2)~temporal speculative drafting constructs candidate future denoising outcomes at multiple depths by filling remaining positions in descending confidence order, all without additional model calls; and (3)~batched verification with hierarchical acceptance evaluates all candidates in a single forward pass and retains the deepest branch whose speculated tokens agree with the verifier's updated predictions, preventing error propagation from aggressive parallel unmasking. These designs allow PSD to combine the gains of both acceleration families within a single decoding pass.

We evaluate PSD on three open-source dLLMs that span different training paradigms across multiple benchmarks. Empirically, PSD achieves a favorable efficiency--quality trade-off across all tested configurations: it delivers additional speedup over temporal-only methods through spatial parallelism, yielding substantially more tokens per forward pass with negligible quality loss, while parallel-only baselines at comparable parallelism levels suffer notable accuracy degradation. Across all model$\times$benchmark settings, PSD consistently matches or improves upon the Pareto frontier established by either family of baselines. Our contributions are summarized as follows:
\begin{itemize}[leftmargin=*]
    \item We identify a complementary relationship between spatial and temporal acceleration in dLLMs: the two axes compress different dimensions of the decoding process, and we empirically show that combining them yields substantially larger speedups than either axis alone.
    
    \item We propose PSD, a training-free, policy-agnostic framework that couples any parallel unmasking policy with speculative verification in a single decoding pass, using a hierarchical acceptance mechanism to prevent error propagation from aggressive parallel unmasking.
    
    \item Experiments on three dLLMs across mathematical reasoning and code generation benchmarks show that PSD achieves up to $5.5\times$ tokens per forward pass with negligible accuracy loss, consistently matching or improving the Pareto frontier of prior methods.

\end{itemize}

\section{Background}
\label{sec:background}
Our framework builds on two lines of work: dLLMs, which formulate text generation as iterative denoising over masked sequences, and speculative decoding, which accelerates inference via a draft-then-verify paradigm. We review each below to motivate the design of PSD.

\subsection{Diffusion Language Models (dLLMs)}

As an emerging paradigm, dLLMs~\citep{DBLP:arxiv/NieZhu25,DBLP:arxiv/YeXie25,DBLP:arxiv/ZhuWang25} formulate text generation as a learned reversal of a token-level masking process. During training, a clean sequence is corrupted by independently replacing tokens with a mask token~$\texttt{[M]}$, and the model is trained to recover the original tokens from the corrupted input \citep{DBLP:arxiv/LiChen25}. At inference, generation starts from a fully masked sequence and proceeds iteratively: at each denoising step, the model predicts tokens for all masked positions and then unmasks a subset of them, progressively refining the sequence until all positions are filled \citep{DBLP:arxiv/BieCao25,DBLP:arxiv/MaDu25,DBLP:arxiv/ZhuYou25}. 

The inference cost is governed by $k$, the number of tokens unmasked per step. When $k\!=\!1$, producing $N$ tokens demands $N$ sequential forward passes, placing dLLMs at a latency regime comparable to autoregressive models of similar size. Setting $k>1$ reduces the total iteration count to $\lceil N/k\rceil$ but forces low-confidence positions to commit prematurely, leading to notable quality degradation~\citep{DBLP:conf/iclr/WuZhang26a,DBLP:arxiv/KongZhang25}. This fundamental tension between throughput and accuracy motivates acceleration methods that can reduce the forward-pass count without sacrificing generation fidelity.

\subsection{Speculative Decoding}
Speculative decoding~\citep{DBLP:conf/icml/leviathankalman23,DBLP:arxiv/chenborgeaud23} accelerates autoregressive LLM inference through a draft-then-verify paradigm. Instead of invoking the large model for every decoding step, a lightweight drafter first proposes several future tokens, and the large target model then evaluates these proposals in a single forward pass \citep{DBLP:conf/acl/xiayang24,DBLP:conf/iclr/SunLi26}. Tokens whose draft predictions agree with the target model are accepted, so one expensive target-model call can advance the decoding process by multiple steps while still preserving the target distribution~\citep{DBLP:conf/naacl/hezhong24,DBLP:conf/acl/luowang25,DBLP:arxiv/oliarojia24,DBLP:arxiv/AbramovichAshkenazi26,DBLP:arxiv/LiWang26}.

Adapting this paradigm to dLLMs, however, is non-trivial~\citep{DBLP:arxiv/AgrawalGarrepalli25,DBLP:arxiv/GaoJi25}. In autoregressive generation, tokens are produced in a fixed left-to-right order, so the next prediction target is always clear. In dLLMs, by contrast, the unmasking order is determined dynamically by model confidence and is not known in advance \citep{DBLP:arxiv/LiMuhtar26}; the unit of speculation therefore shifts from a single next token to an entire set of positions to be revealed at each step.

\begin{figure*}[h]
  \centering 
  \includegraphics[width=0.99\textwidth]{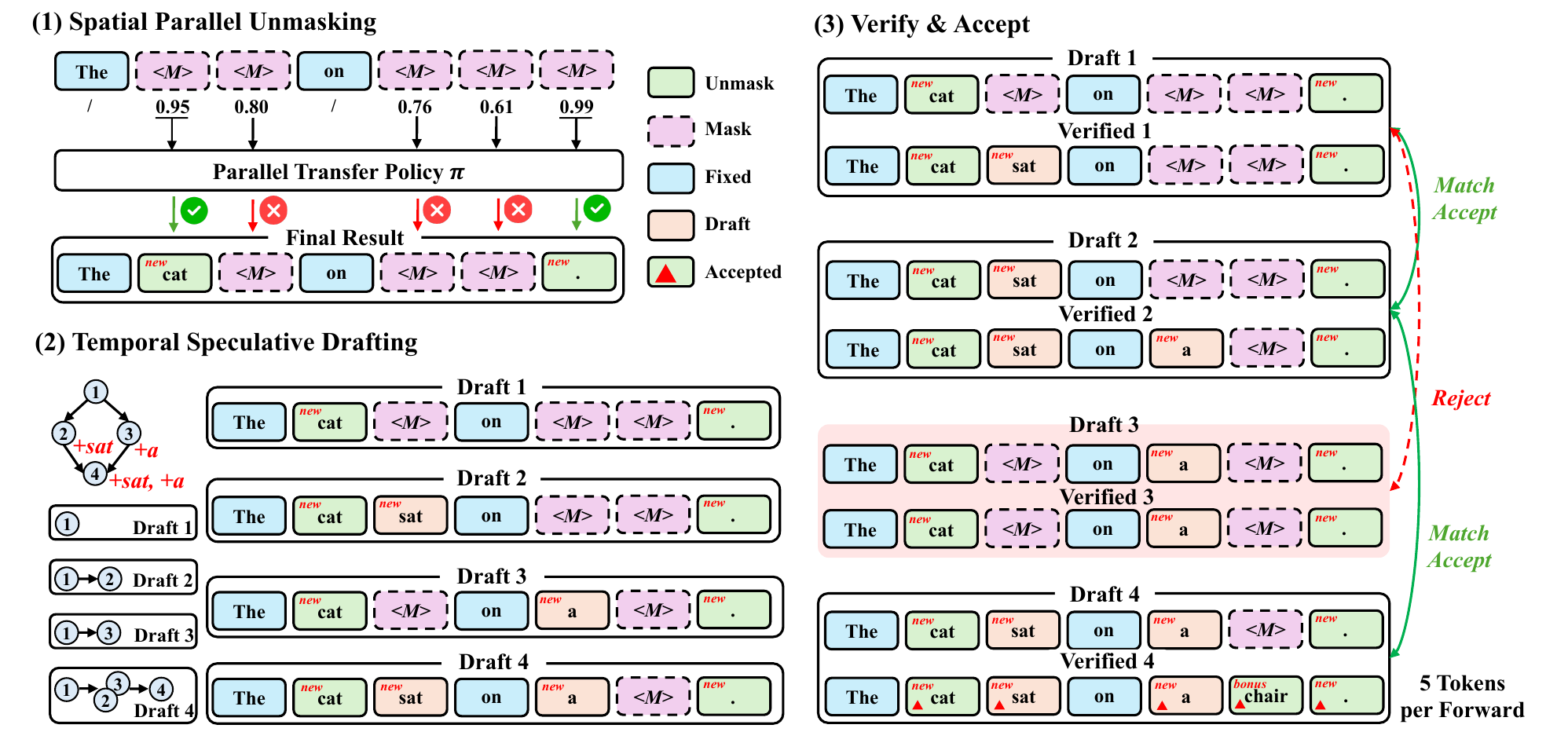}
  \vspace{-0.3em}
    \caption{Overview of the proposed PSD. As a general framework compatible with any parallel decoding strategy, PSD consists of three stages: \textit{(1)~Spatial parallel unmasking} reveals multiple tokens per denoising step; \textit{(2)~Temporal speculative drafting} builds a directed acyclic graph of confidence-ranked drafts for future steps; \textit{(3)~Verify \& Accept} evaluates all drafts in one batched forward pass and accepts the deepest consistent branch. By combining spatial and temporal acceleration, PSD achieves compounded speedups beyond either axis alone.}

  \label{fig:spd_framework}
\end{figure*}

\section{Methodology}


In this section, we first formalize the dLLM decoding process and then present the three stages of PSD (Figure~\ref{fig:spd_framework}): spatial parallel unmasking reveals multiple tokens per step via an arbitrary transfer policy, temporal speculative drafting constructs candidate future outcomes, and batched verification with hierarchical acceptance retains the deepest consistent branch in a single forward pass.

\subsection{Preliminaries}

Denote the sequence state at denoising step $t$ by $\mathbf{x}^{(t)}\in(\mathcal{V}\cup\{\texttt{[M]}\})^N$, where $\mathcal{V}$ is the vocabulary and $\texttt{[M]}$ is a distinguished mask token. At each step $t$ and each masked position $i$, the model assigns a probability $p_\theta(x_i{=}v\mid\mathbf{x}^{(t)})$ to every token $v\in\mathcal{V}$. We denote the most likely token by $\hat{x}_i^{(t)}=\arg\max_{v\in\mathcal{V}} p_\theta(x_i{=}v\mid\mathbf{x}^{(t)})$ and its confidence by $c_i^{(t)}=\max_{v\in\mathcal{V}} p_\theta(x_i{=}v\mid\mathbf{x}^{(t)})$. Let $\mathcal{M}(\mathbf{x}^{(t)})=\{i:x_i^{(t)}=\texttt{[M]}\}$ be the set of masked positions at step $t$. A transfer policy $\pi$ then selects a subset $\mathcal{T}^{(t)}\subseteq\mathcal{M}(\mathbf{x}^{(t)})$ of positions to unmask at that step, and $\bar{r}_\pi=\mathbb{E}[|\mathcal{T}^{(t)}|]$ denotes the average number of positions revealed per step.

\subsection{Parallel Speculative Decoding}
As illustrated in Figure~\ref{fig:spd_framework}, PSD integrates spatial and temporal acceleration into a unified three-stage pipeline. \textit{Stage~(1)~Spatial parallel unmasking} reveals multiple tokens per denoising step via an \emph{arbitrary} transfer policy, PSD is agnostic to this choice. \textit{Stage~(2)~Temporal speculative drafting} reuses the confidence scores from the current step to construct a directed acyclic graph of candidate future denoising outcomes without additional model calls. \textit{Stage~(3)~Batched verification with hierarchical acceptance} evaluates all draft candidates in a single forward pass and accepts the deepest branch that remains consistent with the verifier's updated predictions. We detail each stage below.

\paragraph{Spatial parallel unmasking.} (Stage~(1) in Figure~\ref{fig:spd_framework}.) Although transfer policies such as confidence thresholding can in principle operate over the entire sequence, state-of-the-art dLLMs organize generation into contiguous blocks of length $L$~\citep{DBLP:conf/iclr/WuZhang26a}, a design shared by recent methods including LocalLeap and LoPA~\citep{DBLP:arxiv/YuMa25,DBLP:arxiv/KongZhang25,DBLP:arxiv/XuJin25,DBLP:arxiv/ZhongGu26}. Within each block, all $L$ positions are initially set to $\texttt{[M]}$, and the model then enters an iterative denoising loop indexed by $t = 0, 1, \dots, T$. At iteration $t$, a single forward pass produces the predicted token $\hat{x}_i^{(t)}$ and confidence score $c_i^{(t)}$ for every masked position $i \in \mathcal{M}(\mathbf{x}^{(t)})$. The transfer policy then chooses a set $\mathcal{T}^{(t)}$ of positions to unmask, and the sequence state advances as
\begin{equation}
x_i^{(t+1)}=
\begin{cases}
\hat{x}_{i}^{(t)}, & i\in\mathcal{T}^{(t)},\\[2pt]
x_i^{(t)}, & \text{otherwise}.
\end{cases}
\end{equation}

The loop terminates at step $T$ when $\mathcal{M}(\mathbf{x}^{(T)})=\varnothing$, requiring $T\approx\lceil L/\bar{r}_\pi\rceil$ forward passes. Spatial acceleration alone faces a fundamental ceiling: increasing $\bar{r}_\pi$ reduces $T$ but degrades generation quality, yielding a quality\,/\,speed Pareto frontier that no transfer policy can escape. Importantly, PSD imposes no constraints on the choice of $\pi$: any existing or future parallel decoding policy can serve as the spatial backbone, and the subsequent speculative stages remain unchanged.

\paragraph{Temporal speculative drafting.} (Stage~(2) in Figure~\ref{fig:spd_framework}.) To push beyond this ceiling, PSD couples spatial unmasking with temporal speculation. The key observation is that, after step $t$, the confidence ranking over still-masked positions tends to remain stable across the next few denoising iterations. This ranking can therefore be used to construct draft candidates that approximate future reveal sets without additional model calls. After applying the transfer policy $\pi$ at iteration $t$ to produce $\mathbf{x}^{(t+1)}$, let $m=|\mathcal{M}(\mathbf{x}^{(t+1)})|$ denote the number of positions that remain masked within the block. We reuse the confidence scores from iteration $t$ for these remaining positions and sort them in descending order to obtain the speculative ordering $\boldsymbol{\sigma}=(\sigma_1,\dots,\sigma_m)$, where each $\sigma_j\in\mathcal{M}(\mathbf{x}^{(t+1)})$ and $c_{\sigma_1}^{(t)}\ge\cdots\ge c_{\sigma_m}^{(t)}$. Following the directed draft graph framework of~\citet{DBLP:arxiv/AgrawalGarrepalli25}, we organize $K$ speculative drafts as the nodes of a directed acyclic graph $\mathcal{G}=(\mathcal{N},\mathcal{E})$. Each node $k\in\mathcal{N}$ is associated with a subset $\mathcal{S}_k\subseteq\{\sigma_1,\dots,\sigma_m\}$ of positions to speculatively unmask. The draft sequence $\tilde{\mathbf{x}}_k=(\tilde{x}_{k,1},\dots,\tilde{x}_{k,N})$ is defined entry-wise as
\begin{equation}
\tilde{x}_{k,i}=
\begin{cases}
\hat{x}_{i}^{(t)}, & i\in\mathcal{S}_k,\\[2pt]
x_i^{(t+1)}, & \text{otherwise},
\end{cases}
\end{equation}
which replaces the speculatively filled positions with the cached step-$t$ predictions while leaving already-committed tokens unchanged. The root node $r$ has $\mathcal{S}_r=\varnothing$ (hence $\tilde{\mathbf{x}}_r=\mathbf{x}^{(t+1)}$), representing the current state with no speculation. An edge $(p,k)\in\mathcal{E}$ exists whenever $\mathcal{S}_p\subset\mathcal{S}_k$, indicating that node $k$ extends node $p$ by filling $|\mathcal{S}_k\setminus\mathcal{S}_p|$ additional positions. Because the bidirectional attention of dLLMs makes the unmasking order immaterial, a node may have {multiple parents}, in contrast to the tree structures used in autoregressive speculative decoding, which increases the number of pathways through which a deeper draft can be accepted.

This construction relies on two approximations: (i) the confidence ranking from step $t$ remains a good proxy for the ranking at future steps, and (ii) the predicted tokens $\hat{x}_i^{(t)}$ remain accurate despite context changes from intervening unmaskings. The topology of $\mathcal{G}$ is calibrated offline following~\citet{DBLP:arxiv/AgrawalGarrepalli25} to maximize empirical throughput. At inference time, $\mathcal{G}$ is fixed, and all $K$ drafts are constructed by simple index-level operations on cached predictions, requiring no extra model calls.

\paragraph{Batched verification and hierarchical acceptance.} (Stage~(3) in Figure~\ref{fig:spd_framework}.)
All $K$ draft sequences are concatenated into a single batch and evaluated in a single model forward pass. For each node $k$ and each masked position $i\in\mathcal{M}(\tilde{\mathbf{x}}_k)$, the verifier returns an updated prediction $\hat{x}_{k,i}=\arg\max_{v\in\mathcal{V}} p_\theta(x_i{=}v\mid\tilde{\mathbf{x}}_k)$ and its associated confidence $c_{k,i}=\max_{v\in\mathcal{V}} p_\theta(x_i{=}v\mid\tilde{\mathbf{x}}_k)$. Because all drafts share the same prompt prefix and differ only in the speculatively filled positions, KV-cache sharing can be exploited to reduce redundant computation.

Once all drafts have been verified, the acceptance step traverses $\mathcal{G}$ in topological order to find the deepest draft whose speculated tokens all agree with the verifier's greedy predictions. The root $r$ is unconditionally accepted since it contains no speculation. For each child node $k$ with parent $p$, we check only the newly speculated positions in $\mathcal{S}_k \setminus \mathcal{S}_p$: a speculated token $\tilde{x}_{k,i}$ is accepted if the verifier's prediction at position $i$ satisfies the same acceptance criterion that the transfer policy $\pi$ uses to commit tokens (e.g., the verifier assigns $\tilde{x}_{k,i}$ sufficient confidence). Because $\mathcal{G}$ is a DAG, a child may have multiple parents and is accepted whenever it is consistent with at least one of them. The denoising state then advances to the accepted draft that reveals the most positions, amortizing multiple future unmasking steps into a single verification call. More details are shown in the Appendix \ref{app:details_alg}.

\section{Experiments}\label{sec:experiments}

\begin{figure*}[h]
    \centering
    \includegraphics[width=1\linewidth]{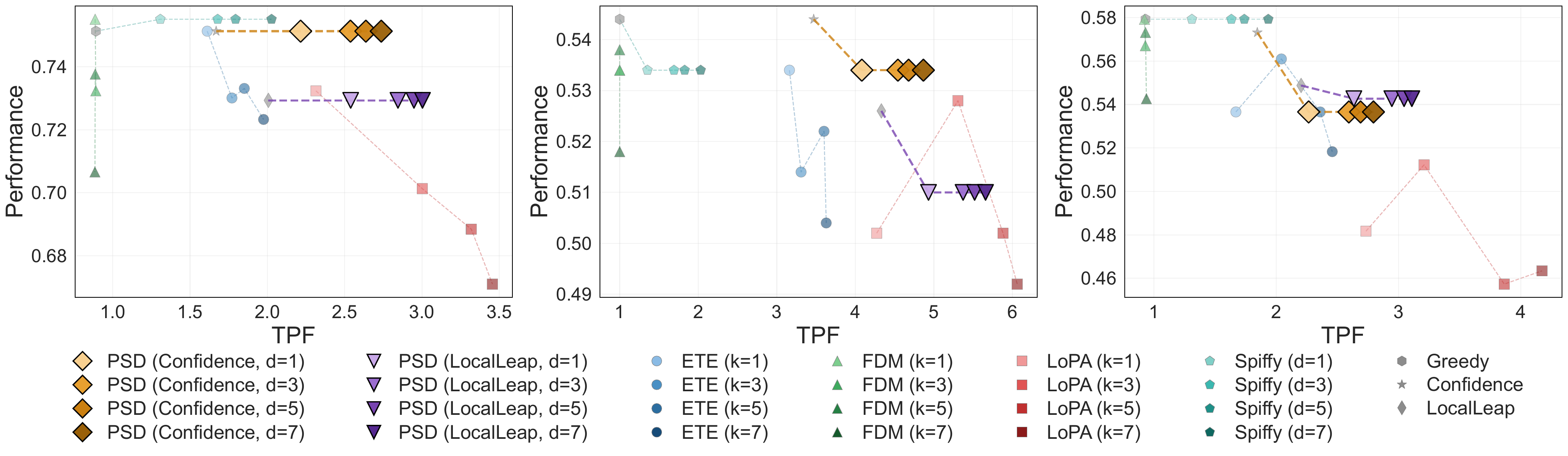}
    \vspace{-1em}
    \caption{Accuracy vs.\ speedup on {Dream-v0-Base-7B} across 27 parameter configurations of different methods. Left: GSM8K; Center: MBPP; Right: HumanEval.}
    \label{fig:dream_scatter}
\end{figure*}

\begin{figure*}[h]
    \centering
    \includegraphics[width=1\linewidth]{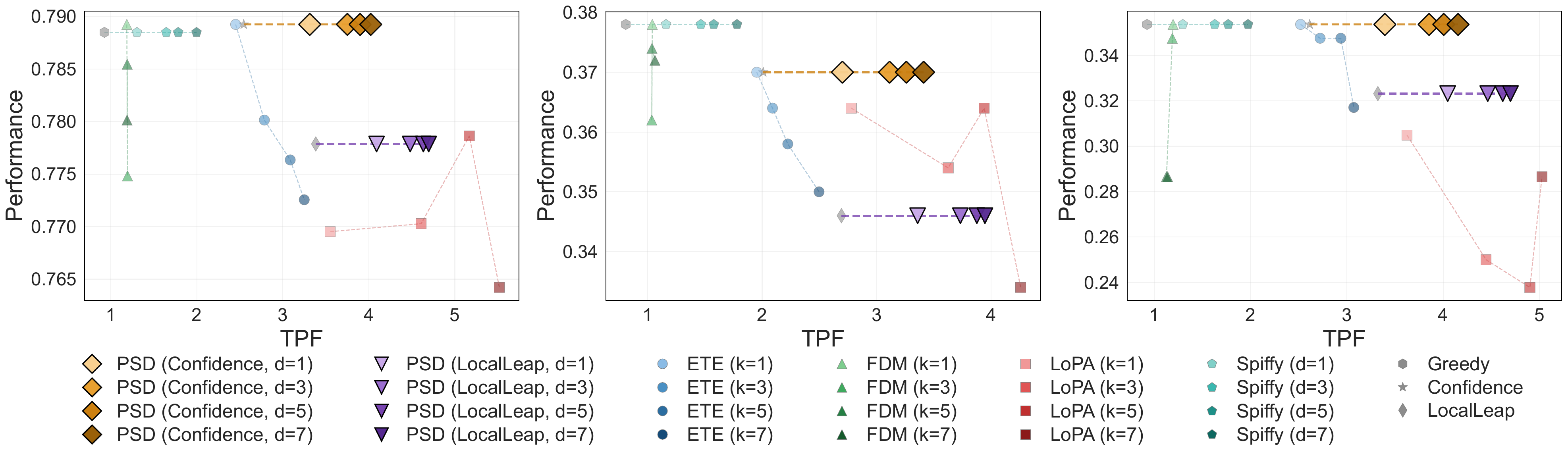}
    \vspace{-1em}
    \caption{Accuracy vs.\ speedup on {LLaDA\,1.5} across 27 parameter configurations of different methods. Left: GSM8K; Center: MBPP; Right: HumanEval.}
    \label{fig:llada_scatter}
\end{figure*}

\begin{figure*}[h]
    \centering
    \includegraphics[width=1\linewidth]{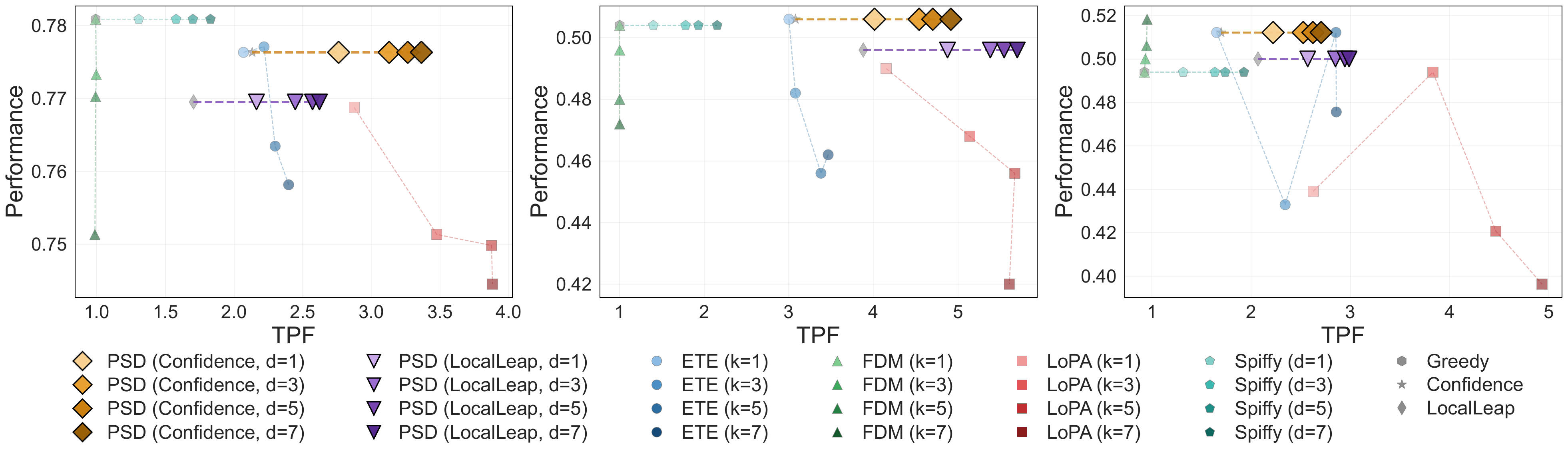}
    \vspace{-1em}
    \caption{Accuracy vs.\ speedup on {openPangu-7B-Diffusion-Base} across 27 parameter configurations of different methods. Left: GSM8K; Center: MBPP; Right: HumanEval.}
    \label{fig:pangu_scatter}
\end{figure*}

In this section, we evaluate PSD on three open-source dLLMs across mathematical reasoning and code generation benchmarks, against seven representative spatial and temporal decoding baselines. Beyond the accuracy-vs-speedup comparison (Figures~\ref{fig:dream_scatter}--\ref{fig:pangu_scatter}), we further analyze the predictability of future decoding positions (Figure~\ref{fig:precision_lookahead}) and the stage-wise contributions of parallel and speculative decoding within a block (Figure~\ref{fig:progress}), which together explain how the two axes jointly contribute to the observed speedups.

\subsection{Experimental Setup}
\paragraph{Models and benchmarks.}
We evaluate three representative open-source dLLMs that cover distinct training paradigms, including AR-initialized fine-tuning, from-scratch diffusion pretraining with preference alignment, and continual diffusion pretraining. Specifically, our model suite consists of Dream-v0-Base-7B~\citep{DBLP:arxiv/YeXie25}, LLaDA-1.5~\citep{DBLP:arxiv/ZhuWang25}, and openPangu-7B-Diffusion-Base\footnote{\url{https://ai.gitcode.com/ascend-tribe/openPangu-7B-Diffusion-Base}}. These models have comparable parameter scales, while differing in their initialization and pretraining strategies, making them suitable for examining whether the observed behaviors are consistent across different dLLM training recipes rather than being tied to a single model family. We conduct experiments on three downstream benchmarks that cover mathematical reasoning and code generation. For mathematical reasoning, we use GSM8K~\citep{DBLP:arxiv/CobbeKosaraju21}, a benchmark of grade-school math word problems requiring multi-step arithmetic reasoning, and report accuracy under chain-of-thought prompting. For code generation, we evaluate on HumanEval~\citep{DBLP:arxiv/ChenTworek21} and MBPP~\citep{DBLP:arxiv/AustinOdena21}, which contain Python programming problems with executable test cases. Further details of the evaluated models and benchmarks are provided in Appendix~\ref{app:additional_models_benchmarks}.




\paragraph{Baseline settings and Metrics.}
We compare PSD against seven representative decoding baselines, including six \textit{spatial} parallel decoding methods and one \textit{temporal} acceleration method; detailed descriptions of all baselines are deferred to the Appendix \ref{app:baseline_details}. Specifically, the spatial baselines include Greedy Decoding, Confidence Parallel Decoding with threshold $\tau{=}0.9$~\citep{DBLP:conf/iclr/WuZhang26a,DBLP:arxiv/YuMa25}, LocalLeap~\citep{DBLP:arxiv/KongZhang25}, LoPA~\citep{DBLP:arxiv/XuJin25}, ETE~\citep{DBLP:arxiv/FuHuang25}, and FDM~\citep{DBLP:journals/corr/abs-2512-04135}, where LoPA, ETE, and FDM each consider a tunable branching or lookahead parameter $k \in \{1,3,5,7\}$. The temporal baseline is Spiffy~\citep{DBLP:arxiv/AgrawalGarrepalli25}, an auto-speculative decoding method with speculative draft depth $d \in \{1,3,5,7\}$. PSD is policy-agnostic and can be combined with arbitrary spatial parallel decoding policies by adding temporal speculation on top. To demonstrate this generality, we instantiate PSD with two representative spatial backbones: {PSD~(Confidence)}, which uses Confidence Parallel Decoding with threshold $\tau{=}0.9$, and {PSD~(LocalLeap)}, which uses LocalLeap with its default parameters. For both PSD variants, we sweep the speculative draft depth $d \in \{1,3,5,7\}$. For task performance, we report accuracy (\%) on GSM8K and pass@1 (\%) on HumanEval and MBPP. For inference efficiency, following~\citet{DBLP:arxiv/AgrawalGarrepalli25}, we report tokens per forward pass (TPF), defined as the average number of tokens decoded per model invocation.

\subsection{Main Results}

\begin{figure}[h]
    \centering
    \includegraphics[width=0.32\linewidth]{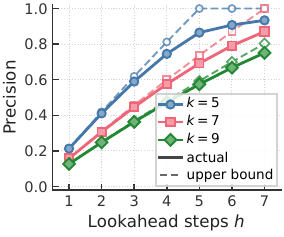}
    \includegraphics[width=0.32\linewidth]{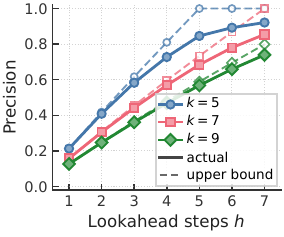}
    \includegraphics[width=0.32\linewidth]{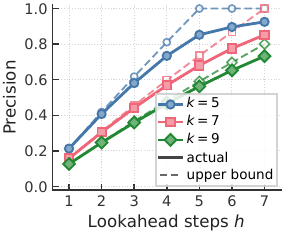}
    \caption{
    Precision@K of the undecoded candidate positions selected at step $t$ for predicting which positions will be decoded within the next $h$ steps on {GSM8K}, {MBPP}, and {HumanEval}. Colors denote different candidate set sizes ($K \in \{5,7,9\}$). Solid lines show empirical precision, and dashed lines show the corresponding oracle upper bound. Notably, this metric demonstrates considerable stability across various experimental settings.
    }
    \label{fig:precision_lookahead}
\end{figure}

\begin{figure}[htbp]
    \centering
    \includegraphics[width=\linewidth]{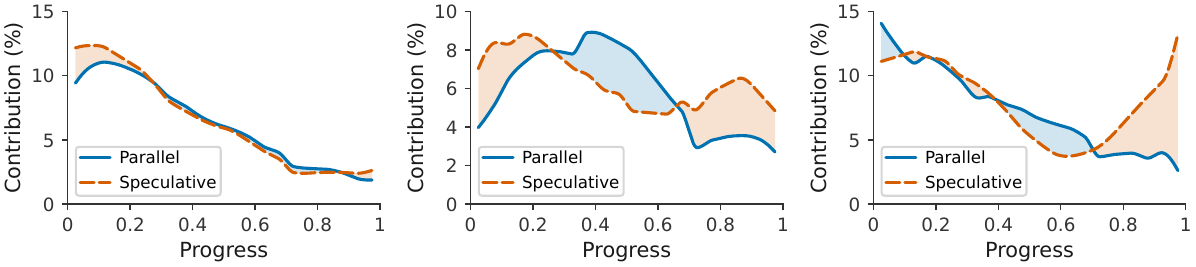}
    \caption{Contribution profiles of parallel decoding and speculative decoding over normalized block progress. The x-axis denotes normalized within-block decoding progress, and the y-axis denotes the percentage of tokens contributed by each mechanism at different progress stages. The three panels correspond to representative blocks from the early, middle, and late phases of generation.}
    \label{fig:progress}
\end{figure}

Figures~\ref{fig:dream_scatter}--\ref{fig:pangu_scatter} plot each method as a point in the accuracy vs.\ speedup plane: rightward indicates higher effective parallelism and upward indicates better task quality. Across all three dLLMs and every benchmark, PSD consistently achieves a superior trade-off between quality and parallelism, occupying the upper-right region of the plane. Furthermore, we also provide two additional analyses in Figures~\ref{fig:precision_lookahead} and~\ref{fig:progress}: the predictability of future decoding positions, and the stage-wise complementarity of parallel and speculative decoding within a block.

\paragraph{Spatial and temporal acceleration combine to yield compounded speedups.}
The defining advantage of PSD is that it compounds two independent sources of parallelism, spatial and temporal, into a single decoding pass, achieving speedups that neither axis can deliver alone. The temporal-only baseline, Spiffy, already improves the greedy baseline's $1.0\times$ parallelism to roughly $1.5$--$2.0\times$ by collapsing multiple denoising iterations via batched verification. PSD further scales this temporal gain by an additional spatial factor: first unmasking multiple tokens per step along the spatial axis and then speculating over future parallel-unmasking steps along the temporal axis. Concretely, on the GSM8K benchmark with LLaDA\,1.5 (Figure~\ref{fig:llada_scatter}, left), PSD~(Confidence) doubles Spiffy's $2.0\times$ to $4.0\times$ tokens per step, contributing an additional $2.0\times$ gain from the spatial axis alone, while preserving 78.9\% accuracy that is virtually identical to greedy decoding (78.8\%). On the MBPP benchmark with Dream (Figure~\ref{fig:dream_scatter}, center), PSD~(Confidence) reaches $4.9\times$ at 53.4\% pass@1, compared with Spiffy's $2.0\times$ at the same accuracy level, yielding a $2.5\times$ additional factor. Similarly, on MBPP with openPangu (Figure~\ref{fig:pangu_scatter}, center), PSD~(Confidence) attains $4.9\times$ while maintaining greedy-level accuracy at 50.6\%. These results suggest that PSD's two-axis design can offer speedups superior to those achievable by any single-axis method alone.

\paragraph{Speculative verification preserves quality under aggressive parallelism.}
Spatial-only baselines face a sharp quality cliff: pushing parallelism higher inevitably forces low-confidence tokens to commit prematurely, causing cascading errors. PSD's hierarchical speculative verification fundamentally mitigates this failure mode. For example, on the HumanEval benchmark with LLaDA\,1.5 (Figure~\ref{fig:llada_scatter}, right), LoPA~$k{=}5$ in its aggressive mode pushes parallelism to $4.9\times$ but collapses pass@1 to 23.8\%, a catastrophic 11.6-point drop from the greedy baseline (35.4\%). In stark contrast, PSD~(Confidence) achieves a comparable parallelism of $4.2\times$ while retaining 35.4\% pass@1 with \emph{zero} degradation. On the GSM8K benchmark with Dream (Figure~\ref{fig:dream_scatter}, left), LoPA~$k{=}7$ in aggressive mode reaches $3.5\times$ but drops accuracy to 67.1\%, losing 8.0 points from the greedy baseline (75.1\%), whereas PSD~(Confidence) at $2.7\times$ maintains the full 75.1\% accuracy without quality loss. The same pattern holds on the MBPP benchmark with openPangu (Figure~\ref{fig:pangu_scatter}, center), where non-speculative methods sacrifice substantial accuracy for marginal additional speedup gains while PSD maintains generation quality at comparable or even higher throughput. Across these settings, PSD generally advances the Pareto frontier of the quality--speedup plane, suggesting that speculative verification is an effective approach for achieving higher parallelism while largely preserving output fidelity.

\paragraph{PSD generalizes across model families and spatial policies.}
Notably, the advantage of PSD appears largely architecture-agnostic. Across Dream (AR-initialized, 7B), LLaDA\,1.5 (from-scratch base, aligned, 8B), and openPangu (continual-pretrained, 7B), PSD generally achieves a favorable quality--speedup trade-off that is competitive with or better than the strongest baselines. Moreover, PSD's modular design exposes a useful speed--quality knob through the choice of spatial policy: PSD~(LocalLeap) tends to achieve higher tokens per step than PSD~(Confidence) on most benchmarks, while PSD~(Confidence) consistently exhibits better accuracy retention (e.g., 75.1\% vs.\ 72.9\% on GSM8K with Dream; 35.4\% vs.\ 32.3\% on HumanEval with LLaDA\,1.5). This plug-and-play flexibility allows practitioners to select a spatial policy suited to their latency or quality constraints, while the temporal speculation layer tends to provide additional gains on top of the chosen policy.

\paragraph{Code generation is more sensitive to spatial parallelism than mathematical reasoning.}
A cross-benchmark comparison reveals that mathematical reasoning (GSM8K) is considerably more tolerant of aggressive parallel unmasking than code generation (HumanEval, MBPP). On the GSM8K benchmark with LLaDA\,1.5 (Figure~\ref{fig:llada_scatter}, left), PSD~(Confidence) at $d{=}7$ reaches $4.0\times$ TPF while maintaining the full greedy-level accuracy of 78.9\%, whereas on the MBPP benchmark with the same model (Figure~\ref{fig:llada_scatter}, center) the same configuration achieves $3.4\times$ but incurs a $0.8$-point drop. This gap widens dramatically for spatial-only baselines: LoPA~$k{=}5$ in aggressive mode loses only $2.0$ points on GSM8K with LLaDA\,1.5 yet collapses by 11.6 points on HumanEval with the same model. The disparity likely stems from tighter inter-token syntactic constraints in code, where a single misplaced delimiter or incorrect identifier propagates fatally through unit-test evaluation, whereas arithmetic reasoning chains can absorb minor token-level perturbations that preserve semantic correctness. Notably, PSD's speculative verification acts as a built-in safety net that helps narrow this sensitivity gap: even on code tasks, PSD~(Confidence) tends to keep the accuracy drop small while still offering substantial speedup. For practitioners, this suggests a practical guideline: code generation workloads may benefit from PSD~(Confidence) with moderate speculative depth to better preserve quality, while math-heavy workloads can more comfortably push toward higher parallelism.

\paragraph{Speculative depth saturates at $d{=}3$ without degrading quality.}
Examining the marginal gains across increasing speculative depths $d \in \{1,3,5,7\}$ reveals a clear diminishing-returns pattern. On the MBPP benchmark with Dream (Figure~\ref{fig:dream_scatter}, center), the non-speculative confidence baseline achieves $3.5\times$ TPF. Adding temporal speculation at $d{=}1$ raises this to $4.1\times$, and $d{=}3$ further extends it to $4.6\times$. Beyond $d{=}3$, however, the curve flattens noticeably, with $d{=}5$ and $d{=}7$ each contributing only an additional $0.1$--$0.2\times$. The three points $d \in \{3,5,7\}$ cluster tightly on the TPF axis. This saturation is expected because the probability of all drafted steps passing hierarchical verification decays geometrically as the speculative horizon lengthens, causing the expected accepted length to converge. Importantly, increasing $d$ does not lead to noticeable accuracy degradation: the PSD points remain largely horizontal in the quality dimension, suggesting that the hierarchical acceptance mechanism is effective at filtering incorrect speculative branches across different depths. This property makes PSD reasonably robust to over-speculation; practitioners may consider $d{=}3$ as a sensible default that captures much of the achievable speedup with modest overhead.

\paragraph{Predictability of Future Decoding Positions.} Figure~\ref{fig:precision_lookahead} shows that the undecoded candidate positions selected at the current decoding step have substantial predictive value for near-future decoding events across all three benchmarks. At small lookahead horizons, Precision@K is consistently high, indicating that a large fraction of the positions selected at time $t$ are indeed decoded within the next few steps. As the lookahead horizon increases, Precision@K gradually decreases, which is expected because the same fixed candidate set must account for a broader future decoding window. Importantly, the precision remains non-trivial over multiple steps, suggesting that the current candidate set captures more than purely instantaneous local information. The gap between the empirical curves and the oracle upper bounds further indicates that the selected candidate set already provides a strong signal about near-future decoding, while still leaving room for improved candidate selection strategies. Overall, these results support the hypothesis that the positions highlighted at the current step can be exploited as effective early indicators of where decoding progress is likely to occur next.

\paragraph{Complementary operating regimes across block phases.}
Across 500 generation trajectories on the GSM8K dataset using LLaDA-1.5, the three panels in Figure~\ref{fig:progress} reveal that parallel and speculative decoding act as complementary axes whose relative dominance shifts with the block's position in generation. In the early-phase block, the two curves decay in tandem, indicating that when context is still being established neither mechanism is preferentially triggered. The middle-phase block exhibits a three-stage structure: speculative leads at the start, parallel peaks in the mid-progress region where accumulated context yields batches of high-confidence tokens, and speculative resurges near the end as remaining masked positions shrink. The late-phase block accentuates this pattern, with parallel dominating the mid-section but speculative rising sharply at completion, as the confidence ordering stabilizes and deeper draft paths are almost always accepted. This stage-wise complementarity is what enables the spatial and temporal axes to interlock rather than compete under PSD.


\section{Related Work}
\label{sec:related_work}

\paragraph{Diffusion Large Language Models (dLLMs).} The sequential generation bottleneck of autoregressive LLMs~\citep{openai-o1,deepseekr1,DBLP:arxiv/AaronAbhimanyu24,DBLP:arxiv/yangli25} has driven growing interest in non-autoregressive alternatives capable of decoding multiple tokens simultaneously. dLLMs~\citep{DBLP:arxiv/NieZhu25,DBLP:arxiv/YeXie25,DBLP:conf/iclr/GongZhang26} address this bottleneck by casting text generation as a discrete denoising process: starting from a fully masked sequence, the model progressively reveals tokens over multiple refinement steps, with bidirectional attention allowing each prediction to condition on the entire sequence context. Scaling this paradigm to billions of parameters has proven effective across both autoregressive-initialized~\citep{DBLP:arxiv/YeXie25} and from-scratch training pipelines with preference optimization~\citep{DBLP:arxiv/ZhuWang25}, producing models that match autoregressive baselines on reasoning and code generation benchmarks~\citep{DBLP:arxiv/MaDu25,DBLP:conf/iclr/WuZhang26a}.

\paragraph{Inference Acceleration for dLLMs.} Training-free acceleration of dLLM inference has been pursued along two largely independent directions. The first reduces denoising iterations by unmasking multiple tokens per step: confidence-based thresholding~\citep{DBLP:conf/iclr/WuZhang26a,DBLP:arxiv/YuMa25}, LocalLeap~\citep{DBLP:arxiv/KongZhang25}, and LoPA~\citep{DBLP:arxiv/XuJin25} each improve per-step parallelism but face an inherent quality ceiling as unmasking grows more aggressive. The second reduces per-iteration cost via speculative decoding. Originally developed for autoregressive LLMs~\citep{DBLP:arxiv/chenborgeaud23,DBLP:conf/icml/leviathankalman23,DBLP:conf/iclr/SunLi26,DBLP:conf/acl/xiayang24} with extensions to tree-structured verification~\citep{DBLP:arxiv/MiaoOliaro23}, multi-head prediction~\citep{DBLP:arxiv/caili24}, and feature-level drafting~\citep{DBLP:conf/icml/liwei24a,DBLP:conf/emnlp/liwei24b,DBLP:arxiv/liwei25}, this paradigm was adapted to dLLMs by Spiffy~\citep{DBLP:arxiv/AgrawalGarrepalli25}, though each step is restricted to a single token. Speculative sampling has also been explored for continuous image diffusion~\citep{DBLP:conf/icml/BortoliGalashov25,DBLP:conf/icml/HuDas25}. PSD operates at the intersection of these two directions, jointly accelerating along both axes.
\section{Conclusion}
\label{sec:conclusion}
We presented PSD, a policy-agnostic framework that composes spatial parallel unmasking with temporal speculative verification to accelerate dLLM inference along both axes simultaneously. The key insight is that confidence rankings over masked positions stabilize across denoising steps, enabling reliable multi-depth draft construction without additional model calls. Because PSD treats the spatial unmasking policy as a modular component, it is compatible with any existing or future parallel decoding method. Experiments across diverse configurations confirm that PSD achieves $3$--$5.5\times$ tokens per forward pass while keeping accuracy within 1 point of greedy decoding on most settings. We believe PSD establishes a general recipe for dLLM acceleration: practitioners can plug in any spatial policy and benefit from temporal speculation on top, opening the door to further gains as stronger parallel decoding methods emerge.

{
\small

\bibliographystyle{unsrtnat}
\bibliography{references}



\clearpage
\appendix

\section{Additional Experimental Setup}
\subsection{Additional Details on Models and Benchmarks}
\label{app:additional_models_benchmarks}
\paragraph{Models.}
We evaluate three open-source dLLMs that are comparable in scale but differ substantially in how diffusion modeling is introduced during training. This selection allows us to examine whether the empirical observations hold across different dLLM construction pipelines, rather than being specific to a single model family or training recipe:
\begin{itemize}[itemsep=1pt, parsep=0pt, topsep=2pt, leftmargin=*]
\item \textbf{Dream-v0-Base-7B}\footnote{\url{https://huggingface.co/Dream-org/Dream-v0-Base-7B}}~\citep{DBLP:arxiv/YeXie25}: a 7B masked diffusion model initialized from Qwen2.5-7B~\citep{DBLP:arxiv/qwen25report} and further pretrained on approximately 580B tokens. We include it as a representative of the \textit{AR-initialized fine-tuning} paradigm, where a pretrained autoregressive backbone is adapted into a diffusion-style language model.
\item \textbf{LLaDA-1.5}\footnote{\url{https://huggingface.co/GSAI-ML/LLaDA-1.5}}~\citep{DBLP:arxiv/ZhuWang25}: an 8B diffusion model pretrained from scratch on 2.3T tokens and further aligned via Variance-Reduced Preference Optimization. We include it as a representative of \textit{from-scratch diffusion pretraining with preference alignment}, where the model is trained natively as a diffusion language model rather than converted from an autoregressive checkpoint.
\item \textbf{openPangu-7B-Diffusion-Base}\footnote{\url{https://ai.gitcode.com/ascend-tribe/openPangu-7B-Diffusion-Base}}: a 7B context-causal block diffusion model initialized from openPangu-Embedded-7B and continually pretrained on 700B tokens. We include it as a representative of \textit{continual diffusion pretraining}. The model uses context-causal block diffusion, where tokens inside the current block are decoded with full attention while previous context is handled with causal attention.
\end{itemize}
Together, these models cover three practically important routes to building dLLMs: adapting an autoregressive model, training a diffusion model from scratch, and continually pretraining an existing model with a block-diffusion objective. This diversity is important for our evaluation because differences in decoding behavior may arise from either the diffusion formulation itself or from the underlying training procedure.

\paragraph{Benchmarks.}
We evaluate on three downstream benchmarks that test two capabilities where generation quality can be judged with relatively objective criteria: mathematical reasoning and code generation. For mathematical reasoning, we use GSM8K; for code generation, we use HumanEval and MBPP:
\begin{itemize}[itemsep=1pt, parsep=0pt, topsep=2pt, leftmargin=*]
\item \textbf{GSM8K}~\citep{DBLP:arxiv/CobbeKosaraju21}: a benchmark of 1,319 grade-school math word problems requiring multi-step arithmetic reasoning. We use chain-of-thought prompting and report accuracy. GSM8K is included to evaluate whether dLLMs can maintain coherent intermediate reasoning steps before producing the final answer.
\item \textbf{HumanEval}~\citep{DBLP:arxiv/ChenTworek21}: a set of 164 hand-crafted Python programming problems, where generated solutions are evaluated by executing hidden unit tests. We report pass@1. HumanEval is included because it requires models to synthesize complete functions that satisfy precise functional specifications.
\item \textbf{MBPP}~\citep{DBLP:arxiv/AustinOdena21}: a benchmark of 500 crowd-sourced Python programming tasks with natural language descriptions and test cases. We report pass@1. Compared with HumanEval, MBPP provides a complementary code-generation setting with shorter and more diverse programming tasks.
\end{itemize}
These benchmarks are suitable for our study because they provide objective and automatically verifiable evaluation signals across reasoning and code-generation tasks. Following the setting in Fast-dLLM~\citep{DBLP:conf/iclr/WuZhang26a}, we use 5-shot prompting for GSM8K, 0-shot prompting for HumanEval, and 3-shot prompting for MBPP. For evaluation, we report flexible-match accuracy on GSM8K to account for minor formatting variations in numerical answers, and pass@1 on HumanEval and MBPP based on unit-test execution.

\section{Additional Implementation Details and Baselines}
\label{app:impl}

\subsection{Implementation Details}
\label{app:impl_details}

All experiments are conducted on a single node equipped with $8\times$ NVIDIA V100 GPUs. We implement our method on top of PyTorch and the HuggingFace Transformers library, and run all baselines within the same codebase and evaluation harness to eliminate discrepancies introduced by differing implementations. Unless otherwise specified, we use a semi-autoregressive block size of $32$ and a maximum generation length of $512$ tokens for all methods. Within each block, tokens are iteratively unmasked according to the policy of the corresponding method, and the next block is only initialized after the current block is fully resolved. All methods share the same tokenizer, prompt templates, and random seeds for a given task so that differences in measured throughput and accuracy are attributable solely to the decoding algorithm. To ensure a fair efficiency comparison, we further enable EOS-based early stopping for every baseline as well as for our method: decoding terminates as soon as an \texttt{<eos>} token is produced. Because parallel decoding methods may unmask tokens \textit{after} the EOS position within the same forward pass, naively counting all emitted tokens would artificially inflate throughput. We therefore compute both the speedup and the tokens-per-forward-pass (TPF) statistics \textit{only over tokens generated up to and including the EOS position}, discarding any tokens unmasked beyond it; the same truncation rule is applied consistently to all methods. 

\subsection{Detailed Baseline Descriptions}
\label{app:baseline_details}

We provide the full descriptions of the seven baselines used in the main paper. All baselines are reimplemented or adapted to use the same backbone model, tokenizer, and block-wise decoding skeleton, so that they differ only in the token-selection rule (spatial axis) or the iteration-compression rule (temporal axis). The first six baselines accelerate decoding along the \textit{spatial} axis by unmasking multiple positions within a single forward pass:
\begin{itemize}[itemsep=2pt, parsep=0pt, topsep=2pt, leftmargin=*]
    \item \textbf{Greedy Decoding}: at each forward pass, unmasks exactly one position---the token with the highest predicted confidence across all currently masked positions in the active block. This method does not exploit any parallelism and therefore serves as the \textit{quality reference}: any deviation of other methods from Greedy's outputs reflects the quality cost of parallel decoding.
    \item \textbf{Confidence Parallel Decoding} ($\tau{=}0.9$)~\citep{DBLP:conf/iclr/WuZhang26a,DBLP:arxiv/YuMa25}: at each forward pass, unmasks \emph{all} positions whose predicted top-1 confidence exceeds a global threshold $\tau$. A higher $\tau$ yields more conservative (quality-preserving) decoding, while a lower $\tau$ unmasks more tokens per step. Following prior work, we fix $\tau{=}0.9$, which provides a strong quality--efficiency operating point.
    \item \textbf{LocalLeap}~\citep{DBLP:arxiv/KongZhang25}: exploits \textit{local determinism propagation}, the empirical observation that tokens neighboring a high-confidence anchor tend to become highly deterministic as well. LocalLeap identifies such anchors and unmasks a local window around each anchor jointly, producing localized parallel decoding without relying on a global confidence threshold. We use the default hyperparameters from the original paper.
    \item \textbf{LoPA}~\citep{DBLP:arxiv/XuJin25}: explores $k$ candidate \textit{Token Filling Orders} at each step and selects the order whose resulting intermediate state has the highest estimated future parallelism potential, thereby decoupling the next unmasking decision from greedy per-token confidence. Here $k$ controls the number of candidate filling orders that are scored at each step; larger $k$ trades extra lookahead computation for better long-horizon parallelism. We sweep $k \in \{1,3,5,7\}$.
    \item \textbf{ETE}~\citep{DBLP:arxiv/FuHuang25}: \textit{strategically explores} high-uncertainty tokens by leveraging \textit{cross-block parallelism}, allowing tokens from adjacent blocks to be unmasked together when their predictions are mutually informative. Here $k$ controls the cross-block exploration width, i.e., how many blocks are jointly considered when selecting positions to unmask in a single forward pass; larger $k$ increases potential parallelism at the cost of greater interference between block predictions. We sweep $k \in \{1,3,5,7\}$.
    \item \textbf{FDM}~\citep{DBLP:journals/corr/abs-2512-04135}: simultaneously predicts the current and future token distributions within a single forward pass, so that tokens at different denoising horizons can be unmasked in parallel. Here $k$ is the number of future steps whose distributions are jointly predicted per forward pass; larger $k$ increases the potential per-step yield but also amplifies prediction drift for distant future tokens. We sweep $k \in \{1,3,5,7\}$.
\end{itemize}
The seventh baseline instead accelerates decoding along the \textit{temporal} axis by compressing the iteration count rather than increasing per-iteration throughput:
\begin{itemize}[itemsep=2pt, parsep=0pt, topsep=2pt, leftmargin=*]
    \item \textbf{Spiffy}~\citep{DBLP:arxiv/AgrawalGarrepalli25}: a \textit{lossless auto-speculative} method that drafts multiple denoising iterations ahead and then verifies them against the target model in a single pass, preserving the exact output distribution of the underlying diffusion decoder. Here $d$ denotes the \textit{speculative draft depth}, i.e., the number of consecutive denoising iterations that are speculatively drafted before each verification step; larger $d$ yields greater potential speedup when drafts are accepted but increases the penalty of rejection. We sweep $d \in \{1,3,5,7\}$.
\end{itemize}

\subsection{Our Method Configurations}
\label{app:our_method_config}

PSD is intentionally designed to be \textit{policy-agnostic}: it treats the spatial parallel decoding rule as a plug-in module and layers temporal speculation on top, so that advances along either axis can be incorporated into PSD without modification. To demonstrate this generality, we instantiate two variants with complementary spatial backbones:
\begin{itemize}[itemsep=2pt, parsep=0pt, topsep=2pt, leftmargin=*]
    \item \textbf{PSD~(Confidence)}: uses Confidence Parallel Decoding with threshold $\tau{=}0.9$ as its spatial backbone. This variant represents a \textit{global, confidence-driven} spatial policy that aggressively unmasks any position the model is already certain about.
    \item \textbf{PSD~(LocalLeap)}: uses LocalLeap with its default parameters as its spatial backbone. This variant represents a \textit{local, structure-driven} spatial policy that exploits determinism propagation around anchors, and does not rely on a fixed global confidence threshold.
\end{itemize}
Both variants share the same temporal speculation module and sweep the speculative draft depth $d \in \{1,3,5,7\}$, where $d$ carries the same semantics as in Spiffy (the number of speculatively drafted iterations per verification step). All other hyperparameters (block size, maximum generation length, tokenizer, and random seeds) are identical to those used for the baselines, so that the reported improvements are attributable purely to PSD's combination of spatial and temporal acceleration.

\section{Pseudocode of Algorithm}
\label{app:details_alg}

\begin{algorithm}[h]
\caption{Parallel Speculative Decoding (PSD)}
\label{alg:psd}
\begin{algorithmic}[1]
\Require dLLM $p_\theta$; transfer policy $\pi$ with acceptance criterion $\mathcal{A}_\pi$; draft DAG $\mathcal{G}=(\mathcal{N},\mathcal{E})$ with root $r$ and node subsets $\{\mathcal{S}_k\}_{k\in\mathcal{N}}$; block length $L$
\Ensure Fully denoised block $\mathbf{x}^\star$
\State Initialize $\mathbf{x}^{(0)} \leftarrow [\texttt{[M]}]^{L}$;\quad $t \leftarrow 0$
\While{$\mathcal{M}(\mathbf{x}^{(t)}) \neq \varnothing$}
    \State \stage{Stage 1: Spatial parallel unmasking}
    \State $\bigl\{(\hat{x}_i^{(t)},\, c_i^{(t)})\bigr\}_{i\in\mathcal{M}(\mathbf{x}^{(t)})} \leftarrow p_\theta(\mathbf{x}^{(t)})$ \Comment{single forward pass}
    \State $\mathcal{T}^{(t)} \leftarrow \pi\bigl(\mathbf{x}^{(t)},\, \{c_i^{(t)}\}\bigr)$ \Comment{positions to commit}
    \State $x_i^{(t+1)} \leftarrow \hat{x}_i^{(t)}$ for $i \in \mathcal{T}^{(t)}$;\quad $x_i^{(t+1)} \leftarrow x_i^{(t)}$ otherwise
    \If{$\mathcal{M}(\mathbf{x}^{(t+1)}) = \varnothing$}
        \State $t \leftarrow t+1$;\quad \textbf{break}
    \EndIf
    \State \stage{Stage 2: Temporal speculative drafting (no extra model call)}
    \State Sort residual masked positions by confidence into $\boldsymbol{\sigma}=(\sigma_1,\dots,\sigma_m)$, $c_{\sigma_1}^{(t)}\!\ge\!\cdots\!\ge\! c_{\sigma_m}^{(t)}$.
    \ForAll{$k \in \mathcal{N}$}
        \State $\tilde{x}_{k,i} \leftarrow \hat{x}_i^{(t)}$ for $i \in \mathcal{S}_k$;\quad $\tilde{x}_{k,i} \leftarrow x_i^{(t+1)}$ otherwise \Comment{index-level assembly}
    \EndFor
    \State \stage{Stage 3: Batched verification with hierarchical acceptance}
    \State $\bigl\{(\hat{x}_{k,i},\, c_{k,i})\bigr\} \leftarrow p_\theta\bigl(\{\tilde{\mathbf{x}}_k\}_{k\in\mathcal{N}}\bigr)$ \Comment{one batched forward pass}
    \State $\mathcal{A} \leftarrow \{r\}$ \Comment{root carries no speculation and is always accepted}
    \ForAll{$k \in \mathcal{N}\setminus\{r\}$ in topological order}
        \If{$\exists\, p \in \mathcal{A}$ with $(p,k)\in\mathcal{E}$ s.t.\ $\mathcal{A}_\pi\bigl(\tilde{x}_{k,i},\hat{x}_{p,i},c_{p,i}\bigr)=\texttt{true}\ \forall i\in\mathcal{S}_k\setminus\mathcal{S}_p$}
            \State $\mathcal{A} \leftarrow \mathcal{A} \cup \{k\}$ \Comment{consistent with at least one accepted parent}
        \EndIf
    \EndFor
    \State $k^\star \leftarrow \arg\max_{k\in\mathcal{A}} |\mathcal{S}_k|$ \Comment{deepest accepted draft}
    \State $\mathbf{x}^{(t+1)} \leftarrow \tilde{\mathbf{x}}_{k^\star}$;\quad additionally commit verifier predictions $\hat{x}_{k^\star,i}$ at positions passing $\mathcal{A}_\pi$
    \State $t \leftarrow t+1$
\EndWhile
\State \Return $\mathbf{x}^\star \leftarrow \mathbf{x}^{(t)}$
\end{algorithmic}
\end{algorithm}

\section{Case Study}
\label{app:casestudy}

To provide a more intuitive understanding of the practical effectiveness of our proposed method, we present a qualitative case study in this section. Specifically, we visualize the generation results produced by PSD (Confidence, $d=7$) and LoPA ($k=7$) on representative examples, allowing for a side-by-side comparison of their output quality. While quantitative metrics reported in previous sections demonstrate the overall performance trends, such case-level visualizations offer deeper insights into how each method behaves during the decoding process. As shown in the following examples, PSD (Confidence, $d=7$) not only achieves a substantial acceleration over standard decoding, but also preserves the fidelity and coherence of the generated content, yielding outputs whose quality remains reliably on par with the baseline.

\section{Limitations}
\label{app:limitations}
While PSD demonstrates strong empirical performance and establishes a favorable quality-efficiency Pareto frontier, our study has certain boundaries that present natural avenues for future exploration. Our empirical analysis is heavily anchored in rigorous mathematical reasoning and code generation benchmarks, which were chosen deliberately for their objective and automatically verifiable correctness. Consequently, evaluating the impact of aggressive spatial-temporal parallel decoding on highly subjective, open-ended generation tasks such as creative writing or long-form summarization remains an area for future investigation. Furthermore, although we demonstrate that default configurations (e.g., a moderate speculative depth) generalize well and capture the majority of the achievable algorithmic speedup across our tested scenarios, deploying PSD in highly specialized or out-of-distribution domains might benefit from further domain-specific hyperparameter tuning to achieve the absolute optimal trade-off between parallel throughput and generation fidelity.

\newpage

\end{document}